\documentclass[sigconf,nonacm]{acmart}

\PassOptionsToPackage{table}{xcolor}

\usepackage{amsmath}
\usepackage{multirow}
\usepackage{threeparttable}
\usepackage{booktabs}
\usepackage{graphicx}
\usepackage{colortbl}

\setcopyright{none}
\renewcommand\footnotetextcopyrightpermission[1]{}
\title{Test-Time Agent Evolution for Long-Horizon Legal Reasoning}

\newcommand{\method}{\textbf{TALA}}

\author{
Haotian Chen$^{1}$,
Shuaicheng Niu$^{2}$,
Haocong Rao$^{2}$,
Kaisong Song$^{3}$,
Jun Lin$^{3}$,
Lizhen Cui$^{1}$,
Zhiqi Shen$^{2}$,
Yonghui Xu$^{1,*}$
}

\affiliation{
  \institution{
  $^{1}$Joint SDU--NTU Centre for Artificial Intelligence Research, Shandong University \\
  $^{2}$Nanyang Technological University \\
  $^{3}$Alibaba Group \\
  \texttt{haotian.chen@mail.sdu.edu.cn, xuyonghui@sdu.edu.cn} \\
  $^{*}$Corresponding author
  }
  \country{\hspace{0pt}}
}
\begin{abstract}
Legal intelligence aims to support reliable decision-making across long-horizon legal processes involving evolving case states and multiple roles.
However, real-world legal deployment exhibits substantial case heterogeneity in facts, evidence, and procedural contexts, exposing the limitations of static agent strategies.
Moreover, legal reasoning is inherently interdependent across roles and procedural stages, making global reliability fundamentally different from isolated role competence.
To address these challenges, we study training-free test-time agent adaptation, where agents continuously exploit deployment-time signals from preceding cases and ongoing interactions without updating model parameters.
We propose \method, which introduces \emph{Test-Time Memory Evolution} to retrieve reusable experience from previous cases, adapt it to the current factual and procedural context, and consolidate accumulated experience for subsequent decision-making.
Further, \emph{Rubric-Aligned Collaboration} verifies and revises role-specific actions according to behavioral and procedural requirements, enabling coordinated decision-making across roles and stages.
Extensive experiments on J1-EVAL and LegalWorld across five backbone models demonstrate consistent improvements over representative reasoning and agent baselines with reasonable interaction and computational costs. Ablation and case studies further show that the two components provide complementary benefits in experience adaptation and cross-role coordination, improving the reliability and efficiency of long-horizon legal reasoning.
\end{abstract}


\begin{document}

\maketitle

\section{Introduction}
\label{sec:intro}







\begin{figure*}[t]
    \centering
    \includegraphics[width=\textwidth]{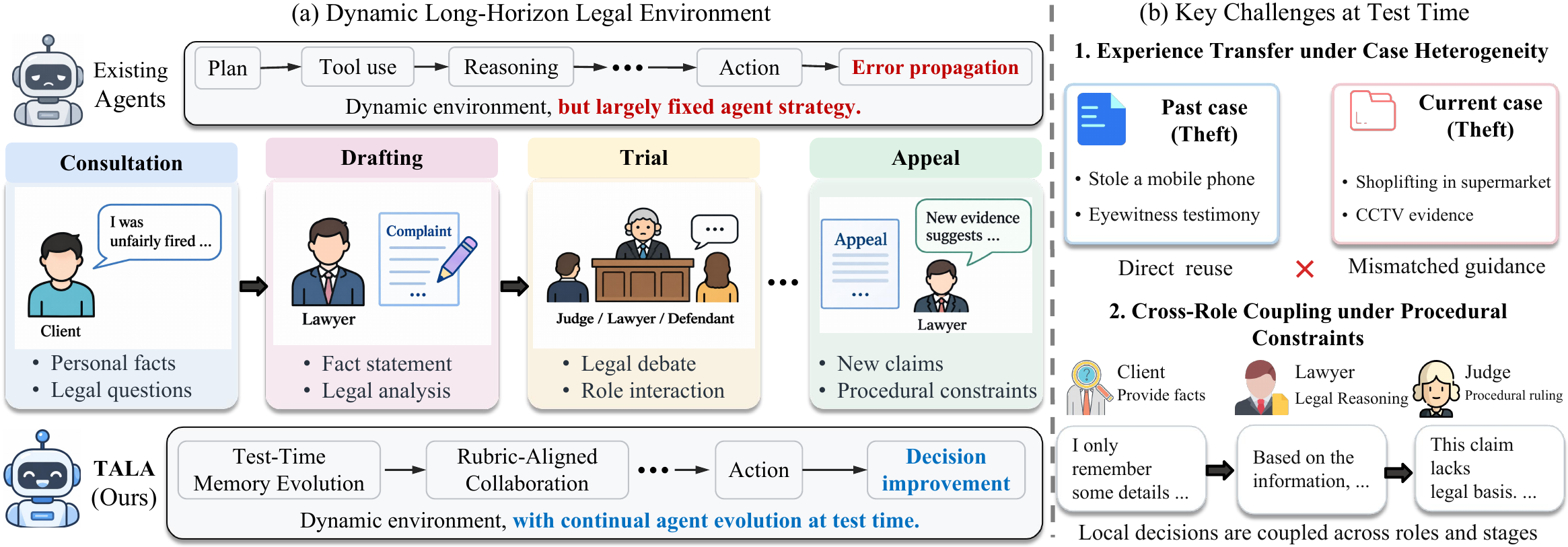}
    \caption{
    (a) Dynamic legal environments evolve across multiple stages, while agent strategies often remain largely fixed during deployment. 
    (b) Case heterogeneity and cross-role misalignment jointly limit effective adaptation in long-horizon reasoning.}
    \label{fig:motivation}
\end{figure*}

Large language models (LLMs) are increasingly deployed as interactive legal agents for legal question answering, document understanding, and agent-based reasoning~\citep{chen2025lexpro,cui2026chatlaw,enguehard2025lemaj,li2025legalagentbench}. However, real-world legal practice rarely consists of isolated tasks or single-turn interactions. Legal consultation, document drafting, trial, and appeal are often sequentially connected, where facts, evidence, claims, and decisions produced at one stage constrain subsequent stages~\citep{zuo2026legalworld}. Legal agents must therefore maintain and update case states while reasoning coherently across extended interactions. The challenge of legal intelligence is thus shifting from merely knowing the law to acting reliably over time.


Recent benchmarks have begun to capture such long-horizon characteristics through multi-turn interaction,  and multi-role legal environments~\citep{jia2026ready,zuo2026legalworld,zhang2026defgen}. \textbf{Nevertheless, a dynamic environment does not necessarily imply an adaptive agent.} Existing agent frameworks typically employ interaction strategies specified before deployment, such as predefined planning, reasoning, and tool-use procedures~\citep{yao2022react,wang2023plan,topsakal2023creating}. Although agents can react to observations within a trajectory, their underlying interaction strategies often remain largely fixed as deployment conditions evolve across cases, roles, and procedural stages. As illustrated in Fig.~\ref{fig:motivation}(a), such static deployment strategies may struggle to adapt to newly emerging information and changing procedural contexts, allowing local errors to propagate into inconsistent arguments and faulty claims. This reveals a fundamental gap between increasingly dynamic legal environments and largely predetermined agent strategies.



In realistic deployment, legal cases arrive continuously, making it impractical to fine-tune or reinforce an agent for every newly encountered case~\citep{cai2025unilaw}. A more practical alternative is to exploit signals naturally available during deployment without updating the underlying model. Recent studies have explored memory- and experience-based mechanisms for reusing information accumulated from previous interactions~\citep{suzgun2026dynamic}. However, experience from one case is not directly transferable to another, as even cases involving the same offense may differ substantially in facts, evidence, and procedural context. As illustrated in Fig.~\ref{fig:motivation}(b1), naively reusing prior experience can therefore introduce mismatched or misleading guidance. \textit{The first challenge is thus how to reconcile cross-case experience transfer with case-specific adaptation across heterogeneous legal cases.}

On the other hand, long-horizon legal reasoning is inherently coupled across roles and procedural stages. Existing methods often improve individual legal roles in isolation~\citep{yue2025multi,wu2024knowledge,yang2026lawthinker}, yet stronger local decisions do not necessarily lead to reliable end-to-end execution. As illustrated in Fig.~\ref{fig:motivation}(b2), decisions made by one role shape the context for subsequent roles, making cross-role misalignment particularly harmful as local inconsistencies can propagate across stages and compromise end-to-end reliability. \textit{The second challenge is therefore how to jointly coordinate decisions across coupled legal roles and procedural stages toward globally consistent execution.}

To address these challenges, we aim to enable LLM agents to adapt throughout long-horizon legal interactions using signals naturally available during deployment. We propose \method, a training-free framework that improves agent behavior at test time without updating the underlying model parameters. To handle case heterogeneity, \method ~introduces \textit{Test-Time Memory Evolution (TME)}, which retrieves relevant experience from preceding cases, adapts it to the current factual and procedural context, and consolidates accumulated experience under a bounded memory budget. This allows prior interactions to provide reusable guidance without directly transferring case-specific facts or evidence. To mitigate cross-role misalignment, we further introduce \textit{Rubric-Aligned Collaboration (RAC)}, which verifies and revises each role's actions according to role-specific behavioral and procedural requirements, preventing locally plausible decisions from introducing inconsistencies into subsequent roles and stages. Together, these mechanisms enable \method{} to evolve both cross-case experience and cross-role decision-making throughout continual legal deployment, improving the global reliability of long-horizon interactions.

Extensive experiments on J1-EVAL and LegalWorld demonstrate that \method{} consistently improves performance across five backbone models and diverse long-horizon legal scenarios, showing strong generalization across model families and scales. More importantly, \method{} substantially improves the robustness of multi-stage execution, sustaining reliable interaction across extended trajectories while maintaining moderate computational and interaction costs. Ablation and case studies further reveal that the two components are complementary and mutually reinforcing, with their joint use yielding more coherent long-horizon adaptation than either mechanism in isolation.

Our contributions can be summarized as follows. 
(1) We formulate long-horizon legal reasoning from a deployment-time evolution perspective, highlighting two fundamental challenges that arise under continual legal deployment: \emph{case heterogeneity}, which complicates cross-case experience transfer, and \emph{cross-role misalignment}, which makes local decisions consequential for subsequent roles and stages. 
(2) We propose \method, a training-free framework for continual test-time agent evolution. It combines \emph{TME} for context-aware cross-case experience adaptation with \emph{RAC} for coordinated decision-making across coupled legal roles and procedural stages. 
(3) Extensive experiments on J1-EVAL and LegalWorld across five backbone models demonstrate consistent gains across model families and scales, improved robustness in long-horizon execution, and moderate test-time overhead. 

\section{Related Work}

\textbf{LLM-based Legal Agents}.
Early studies on legal LLMs mainly focus on static legal capabilities, such as legal knowledge acquisition, judgment prediction, and document understanding, through benchmark construction and domain-specific adaptation~\citep{fei2024lawbench,yue2023disc,fei2025internlm,colombo2024saullm}. More recently, research has shifted toward LLM-based legal agents that support interaction, tool use, and multi-role collaboration in complex legal environments~\citep{cui2026chatlaw,chen2025agentcourt}. Existing methods further enhance legal agents through predefined reasoning workflows and legal-specific tools~\citep{yang2026lawthinker}, or additional training such as supervised fine-tuning and reinforcement learning~\citep{cai2025unilaw}. However, these improvements are largely determined before deployment, while many case-specific facts, procedural states, role interactions, and tool feedback only emerge during execution. \textit{How to exploit such deployment-time signals to continuously adapt subsequent agent behavior remains underexplored.} In this paper, we aim to develop self-improving legal agents for more reliable reasoning.

\textbf{Test-Time Adaptation}.
Test-time adaptation aims to improve model performance during inference without accessing the original training data~\citep{chen2026test}. A prominent line of inference-time improvement is test-time scaling, which allocates additional computation through repeated sampling, majority voting, self-consistency, or iterative refinement~\citep{wang2022self,madaan2023self,snell2025scaling}. While effective for short-form reasoning, such strategies become increasingly costly in long-horizon tasks, where a single trajectory may involve numerous interaction turns, role transitions, and tool calls~\citep{huang2025best,huang2026sample}. More fundamentally, long-horizon agent interaction is inherently non-stationary: each action may alter subsequent states, available information, and behavioral constraints, making future decisions dependent on earlier interactions. \textit{Consequently, both useful experience and local errors can persist and compound over time.} However, existing test-time methods have largely focused on isolated reasoning instances, leaving continual adaptation under interaction-induced non-stationarity relatively underexplored. This paper aims to bridge this gap through test-time agent evolution in long-horizon interactions.
\section{Method}
\label{sec:method}
\begin{figure*}[t]
    \centering
    \includegraphics[width=0.95\textwidth]{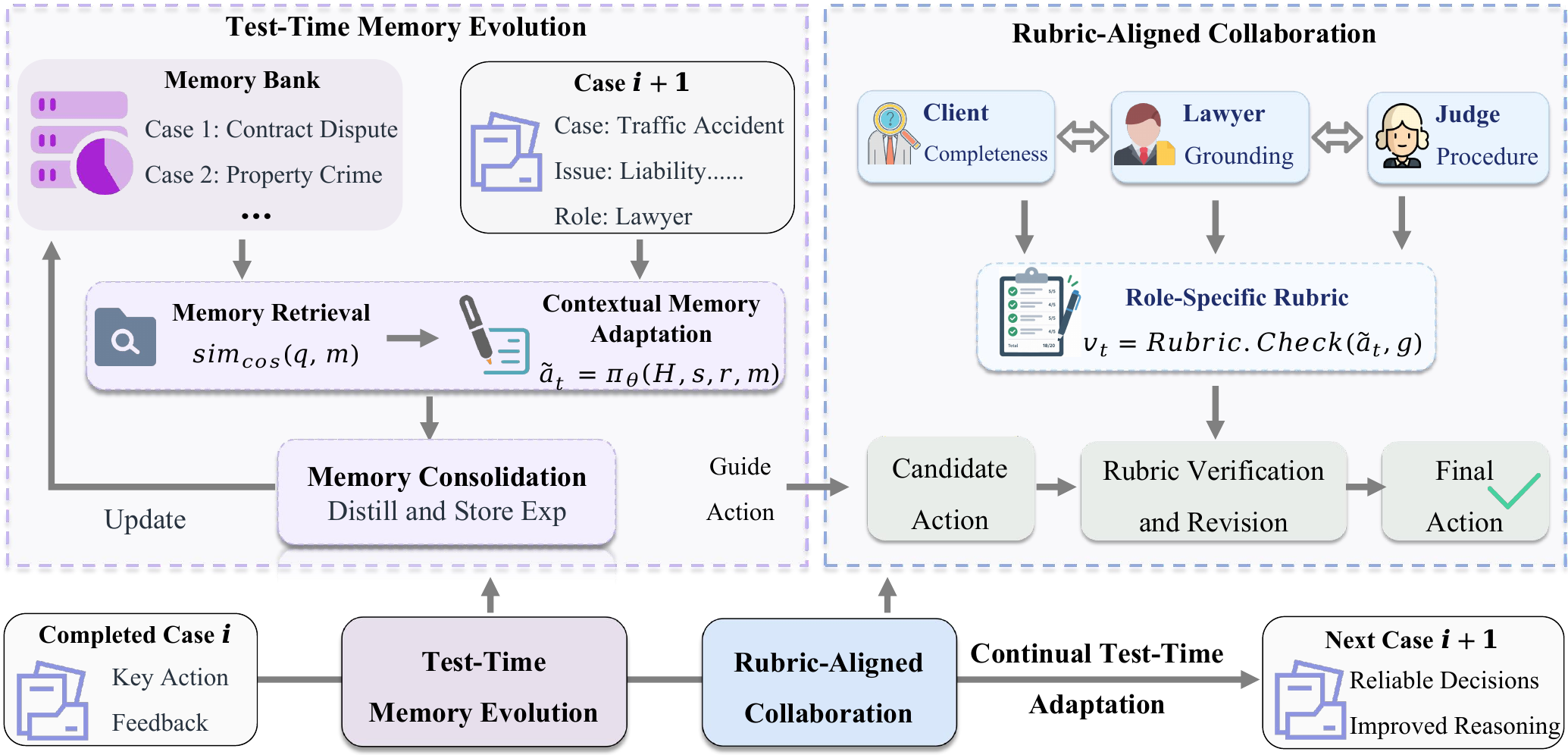}
    \caption{Overall framework of \method. 
    Under a continual test-time setting, the agent sequentially interacts with evolving legal cases and leverages deployment-time signals without updating model parameters. 
    \method~~consists of two components, \textit{Test-Time Memory Evolution} and \textit{Rubric-Aligned Collaboration}, which adapt prior experience to the current case and align role-specific behaviors to improve subsequent decisions.}
    \label{fig:framework}
\end{figure*}

As shown in Fig.~\ref{fig:framework}, \method{} operates under a continual test-time setting, where legal cases are processed sequentially without updating the underlying model parameters. It consists of two complementary components: \emph{Test-Time Memory Evolution}, which retrieves, adapts, and consolidates experience from preceding cases to handle case heterogeneity, and \emph{Rubric-Aligned Collaboration}, which verifies and revises role-specific actions under behavioral and procedural constraints. Together, the two components enable the agent to continuously exploit deployment-time signals for more reliable long-horizon legal reasoning.


\subsection{Problem Formulation}

\textbf{Test-Time Evolution}. Let $\pi_{\Theta}$ denote a pretrained language agent with parameters $\Theta$, learned from training data
$\mathcal{D}_{\mathrm{train}} \sim \mathcal{P}_{\mathrm{train}}$.
At deployment, the agent sequentially encounters domain-specific legal tasks drawn from
$\mathcal{P}_{\mathrm{test}}$, where
$\mathcal{P}_{\mathrm{test}} \neq \mathcal{P}_{\mathrm{train}}$.
As deployment progresses, newly encountered cases may differ in facts, evidence, procedural states, and role interactions, requiring the agent to continually adapt its behavior to evolving test-time conditions.

\textbf{Long-Horizon Interaction}. We formulate each legal case $i$ as a multi-turn interaction involving a set of roles
$\mathcal{R}=\{r_1,\ldots,r_K\}$, such as the lawyer, client, judge, and other legal participants.
Within case $i$, $t$ indexes the interaction turn, and $T_i$ denotes the total number of turns in the corresponding trajectory.
For clarity, we omit the case index $i$ when referring to variables within a single case.
At turn $t$, let $\mathcal{H}_t$ denote the accumulated interaction history, including previous utterances and tool observations, and let $s_t$ denote the current task state, such as the active scenario or litigation stage, procedural progress, and collected case information.
Given the active role $r_t$, the agent produces an action:
\begin{equation}
    a_t =
    \pi_{\Theta}
    \left(
    \mathcal{H}_t, s_t, r_t
    \right),
\end{equation}
where $a_t \in \mathcal{A}$ and $\mathcal{A}$ denotes the action space.
An action can be a natural-language response, such as questioning, answering, or document generation, or an executable action, such as legal retrieval or other tool invocation.
After executing $a_t$, the environment returns an observation $o_{t+1}$ and updates the task state from $s_t$ to $s_{t+1}$.
The trajectory of case $i$ is therefore represented as:
\begin{equation}
    \tau^{(i)} =
    \left\{
    (s_t, r_t, \mathcal{H}_t, a_t, o_{t+1})
    \right\}_{t=1}^{T_i}.
\end{equation}

The agent sequentially processes
$N$ legal cases
$\{\tau^{(i)}\}_{i=1}^{N}$ drawn from $\mathcal{P}_{\mathrm{test}}$ during test time.
Our goal is to exploit information accumulated from preceding cases and their interactions to improve performance on subsequent cases, while maintaining coherent behavior with respect to the current role, task state, and procedure.

\subsection{Test-Time Memory Evolution}
Real-world legal tasks involve complex, multi-stage decision-making~\citep{liu2026llm}, often requiring external legal knowledge and specialized tools. However, existing LLM agents remain unreliable in tool use and long-horizon interactions~\citep{patil2025berkeley}. To address this, we propose a \textit{Test-Time Memory Evolution (TME)} mechanism that continuously accumulates and refines interaction experience during inference to support subsequent decision-making.

The mechanism consists of three components: \emph{Memory Retrieval}, \emph{Contextual Memory Adaptation}, and \emph{Memory Consolidation}. 
After completing each legal case, the agent preserves reusable experience from its interaction trajectory, such as effective tool-use strategies and action guidance, to support subsequent cases. 
Let $\mathcal{M}_i=\{m_j\}_{j=1}^{|\mathcal{M}_i|}$ denote the memory bank available before processing case $i$, where each $m_j$ represents a textual experience entry with its associated metadata.

\emph{Memory Retrieval}. At turn $t$ of case $i$, we retrieve relevant experience from $\mathcal{M}_i$ according to the current interaction context. 
Specifically, we construct a retrieval query $q_t^{(i)}$ from the current task information and encode both the query and memory entries using a BGE encoder $f_{\mathrm{BGE}}(\cdot)$~\citep{bge_embedding}. 
The relevance between $q_t^{(i)}$ and a memory entry $m_j$ is measured by cosine similarity:
\begin{equation}
    \rho_j =
    \operatorname{sim}_{\cos}
    \left(
        f_{\mathrm{BGE}}(q_t^{(i)}),
        f_{\mathrm{BGE}}(m_j)
    \right),
\end{equation}
where $f_{\mathrm{BGE}}(\cdot)$ denotes the BGE embedding encoder and $\rho_j$ measures the semantic similarity between the current query and memory entry $m_j$. We then rank all memory entries according to $\rho_j$ and retain at most the top-$k$ entries whose similarity scores exceed a threshold $\delta$:
\begin{equation}
    \mathcal{R}_t^{(i)}
    =
    \operatorname{TopK}_{k}
    \left(
        \left\{
        (m_j,\rho_j)
        \mid
        m_j \in \mathcal{M}_i,\;
        \rho_j \geq \delta
        \right\}
    \right),
\end{equation}
where $\delta=0.5$ in our implementation. 
$\mathcal{R}_t^{(i)}$ contains the retrieved textual experience entries and their metadata, while the embeddings are used only for similarity computation.

\emph{Contextual Memory Adaptation.}
Although retrieved memories may originate from the same legal scenario, their case-specific facts and evidentiary conditions can differ substantially. For example, cases involving the same offense may concern different objects, circumstances, or supporting evidence. Directly reusing such memories may therefore introduce irrelevant or mismatched guidance. We use the same agent $\pi_{\Theta}$ to adapt the retrieved experience to the current interaction context:
\begin{equation}
\begin{aligned}
    \hat{m}_t^{(i)}
    &=
    \pi_{\Theta}
    \left(
        \mathcal{R}_t^{(i)},
        \mathcal{H}_t^{(i)},
        s_t^{(i)}
    \right), \\
    \tilde{a}_t^{(i)}
    &=
    \pi_{\Theta}
    \left(
        \mathcal{H}_t^{(i)},
        s_t^{(i)},
        r_t^{(i)},
        \hat{m}_t^{(i)}
    \right),
\end{aligned}
\end{equation}
where $\hat{m}_t^{(i)}$ denotes the context-adapted memory incorporated into the current interaction context, and $\tilde{a}_t^{(i)}$ denotes the preliminary action generated before rubric-based verification.

\emph{Memory Consolidation.}
Instead of directly storing raw interaction histories, we first distill each completed interaction into a concise and reusable memory entry that summarizes effective retrieval strategies and action guidance. The distilled entry is then encoded by $f_{\mathrm{BGE}}(\cdot)$ and added to the persistent memory bank.

As the memory bank grows, redundant or noisy experiences may reduce retrieval precision rather than further improve reasoning~\citep{cao2026higmem}. We therefore maintain a bounded memory bank and trigger global pairwise consolidation once its size exceeds a predefined capacity $C$. Specifically, we select the most semantically similar pair $(p^{*},q^{*})=\arg\max_{p<q}\operatorname{sim}_{\cos}\!\left(f_{\mathrm{BGE}}(m_p),f_{\mathrm{BGE}}(m_q)\right)$ over the entire memory bank, rather than restricting the comparison to pairs that involve the newly added memory. The same agent $\pi_{\Theta}$ then merges the selected pair into a consolidated experience $m_{p,q}^{*}=\pi_{\Theta}(m_{p^{*}},m_{q^{*}})$, and updates the memory bank as:
\begin{equation}
    \mathcal{M}
    \leftarrow
    \left(
        \mathcal{M}
        \setminus
        \{m_{p^{*}},m_{q^{*}}\}
    \right)
    \cup
    \{m_{p,q}^{*}\}.
\end{equation}
This consolidation removes redundant experiences while preserving reusable knowledge within a fixed memory budget. If merging fails, the oldest memory is removed as a fallback.

\subsection{Rubric-Aligned Collaboration}

We introduce \emph{Rubric-Aligned Collaboration (RAC)} to coordinate multiple roles under scenario-specific rubrics. Rather than optimizing each role independently~\citep{yue2025multi,wu2024knowledge}, each role follows its own interaction constraints, enabling complementary behaviors for reliable legal reasoning and task completion.

For each scenario, we define a set of role-specific rubrics
$\mathcal{G}=\{\mathcal{G}_{\mathrm{lawyer}},
\mathcal{G}_{\mathrm{public}},
\mathcal{G}_{\mathrm{judge}},\ldots\}$,
where each rubric specifies behavioral and procedural requirements for the corresponding role.
For the active role $r_t^{(i)}$, we denote its rubric as
$\mathcal{G}_{r_t^{(i)}}=\{g_1,\ldots,g_J\}$.
At turn $t$, the candidate action $\tilde{a}_t^{(i)}$ is checked against the rubric:
\begin{equation}
    v_t^{(i)}
    =
    \prod_{j=1}^{J}
    v_{t,j}^{(i)},
    \quad
    v_{t,j}^{(i)}
    =
    \operatorname{RubricCheck}
    \left(
        \tilde{a}_t^{(i)}, g_j
    \right)
    \in \{0,1\}.
\end{equation}
Here, $v_{t,j}^{(i)}=1$ indicates that the candidate action satisfies rubric item $g_j$.
Thus, $v_t^{(i)}=1$ only when all rubric requirements are satisfied; otherwise, the violated rubric items are used to provide revision guidance for the candidate action.

The candidate action is directly accepted if it satisfies all rubric requirements; otherwise, the same agent revises it according to the corresponding rubric:
\begin{equation}
    a_t^{(i)}
    =
    \begin{cases}
        \tilde{a}_t^{(i)},
        & v_t^{(i)} = 1, \\[1mm]
        \begin{aligned}
        \pi_{\Theta}\!\big(
            &\mathcal{H}_t^{(i)}, s_t^{(i)}, r_t^{(i)}, \\
            &\hat{m}_t^{(i)}, \tilde{a}_t^{(i)},
            \mathcal{G}_{r_t^{(i)}}
        \big),
        \end{aligned}
        & v_t^{(i)} = 0.
    \end{cases}
\end{equation}
Here, $a_t^{(i)}$ denotes the final action after rubric verification and possible revision. This verification is applied to all participating roles throughout the interaction trajectory, preventing local violations from propagating across turns and thereby reducing error accumulation in long-horizon interactions.
\section{Experiment Setup}
\label{sec:experiments}

\begin{table*}[!t]
\centering
\caption{Performance comparisons on J1-EVAL using Qwen3.5-4B, Qwen3-8B, and Qwen3-32B backbones. Best results are in \textbf{bold} and second-best results are \underline{underlined}.}
\label{tab:main_result_J1Eval}

\resizebox{0.9\linewidth}{!}{
\begin{threeparttable}
\begin{tabular}{lccccccccccccccc}
\toprule

\multirow{3}{*}{\textbf{Method}}
& \multicolumn{2}{c}{\textbf{Level-I}}
& \multicolumn{4}{c}{\textbf{Level-II}}
& \multicolumn{9}{c}{\textbf{Level-III}} \\

\cmidrule(lr){2-3}
\cmidrule(lr){4-7}
\cmidrule(lr){8-16}

& \textbf{KQ}
& \textbf{LC}
& \multicolumn{2}{c}{\textbf{CD}}
& \multicolumn{2}{c}{\textbf{DD}}
& \multicolumn{4}{c}{\textbf{CI}}
& \multicolumn{5}{c}{\textbf{CR}} \\

\cmidrule(lr){2-2}
\cmidrule(lr){3-3}
\cmidrule(lr){4-5}
\cmidrule(lr){6-7}
\cmidrule(lr){8-11}
\cmidrule(lr){12-16}

& \textbf{Avg.}
& \textbf{Avg.}
& \textbf{FOR}
& \textbf{DOC}
& \textbf{FOR}
& \textbf{DOC}
& \textbf{PFS}
& \textbf{JUD}
& \textbf{REA}
& \textbf{LAW}
& \textbf{PFS}
& \textbf{CRI}
& \textbf{VER}
& \textbf{REA}
& \textbf{LAW} \\

\midrule[1.0pt]

\textit{\textbf{Qwen3.5-4B}}
& \underline{44.5}
& 26.2
& 4.3
& 28.2
& 1.4
& 17.3
& 27.8
& 6.3
& 9.1
& 8.7
& \underline{16.4}
& 53.6
& 36.6
& \underline{18.8}
& \underline{25.4} \\

\hspace{6pt} $\bullet$ ReAct
& 25.0
& 18.3
& 4.3
& 24.1
& 0.4
& 10.1
& 1.1
& 4.5
& 0.8
& 9.2
& 0.4
& 5.8
& 4.3
& 1.7
& 2.2 \\

\hspace{6pt} $\bullet$ Plan-and-Solve
& 21.2
& 12.2
& 3.2
& 18.8
& 4.5
& 12.4
& 3.8
& 2.2
& 5.4
& 5.9
& 3.0
& 24.6
& 17.0
& 6.4
& 6.5 \\

\hspace{6pt} $\bullet$ Plan-and-Execute
& 33.4
& 21.9
& 8.6
& 14.1
& 10.2
& 5.0
& 5.4
& 2.9
& 4.4
& 2.4
& 2.7
& 20.3
& 13.0
& 4.8
& 7.4 \\

\hspace{6pt} $\bullet$ LawThinker
& 43.3
& \underline{28.6}
& \underline{53.8}
& \underline{32.5}
& \underline{30.1}
& \underline{21.3}
& \underline{39.8}
& \underline{8.5}
& 9.8
& \underline{12.3}
& 13.2
& 44.9
& \underline{40.7}
& 14.1
& 22.7 \\

\hspace{6pt} $\bullet$ Reflexion
& 25.3
& 15.7
& 0.6
& 17.8
& 1.8
& 8.8
& 1.3
& 0.8
& \underline{10.6}
& 6.8
& 7.5
& \underline{54.3}
& 3.3
& 14.6
& 10.4 \\

\rowcolor{gray!12}
\hspace{6pt} $\bullet$ \textbf{\method}
& \textbf{47.9}
& \textbf{34.2}
& \textbf{69.8}
& \textbf{50.0}
& \textbf{56.6}
& \textbf{42.6}
& \textbf{56.2}
& \textbf{14.0}
& \textbf{21.1}
& \textbf{22.7}
& \textbf{45.3}
& \textbf{64.2}
& \textbf{60.6}
& \textbf{25.3}
& \textbf{42.2} \\

\midrule[1.0pt]

\textit{\textbf{Qwen3-8B}}
& 48.3
& 26.7
& 7.5
& \underline{58.1}
& 8.2
& \underline{35.9}
& 5.2
& 2.9
& 4.3
& 4.7
& \underline{13.6}
& \underline{69.6}
& 48.9
& 30.4
& \underline{40.0} \\

\hspace{6pt} $\bullet$ ReAct
& \textbf{59.5}
& 34.4
& 24.1
& 53.8
& 18.6
& 28.7
& 1.2
& 0.6
& 2.2
& 1.6
& 6.7
& 65.2
& 48.1
& \underline{31.0}
& 34.5 \\

\hspace{6pt} $\bullet$ Plan-and-Solve
& 57.0
& \underline{34.6}
& 20.0
& 38.6
& 6.5
& 8.1
& 1.8
& 2.5
& 5.8
& 5.0
& 0.8
& 66.7
& 46.1
& 25.9
& 25.6 \\

\hspace{6pt} $\bullet$ Plan-and-Execute
& 57.7
& \underline{34.6}
& 8.6
& 22.3
& 0.7
& 3.3
& 7.3
& 5.2
& 6.7
& 5.9
& 7.3
& 56.5
& 31.9
& 19.9
& 21.2 \\

\hspace{6pt} $\bullet$ LawThinker
& \underline{57.9}
& 30.3
& \underline{89.2}
& 52.2
& \textbf{78.9}
& 33.4
& \underline{17.0}
& \underline{8.0}
& \underline{10.3}
& \underline{8.9}
& 11.3
& 60.9
& \underline{55.9}
& 25.8
& 34.4 \\

\hspace{6pt} $\bullet$ Reflexion
& 57.6
& 33.7
& 32.7
& 56.4
& 16.8
& 25.9
& 1.1
& 0.5
& 1.6
& 0.5
& 7.8
& 65.2
& 39.1
& 29.6
& 30.1 \\

\rowcolor{gray!12}
\hspace{6pt} $\bullet$ \textbf{\method}
& 57.5
& \textbf{36.5}
& \textbf{98.0}
& \textbf{75.5}
& \underline{58.7}
& \textbf{55.4}
& \textbf{40.0}
& \textbf{23.5}
& \textbf{29.0}
& \textbf{33.6}
& \textbf{80.0}
& \textbf{97.1}
& \textbf{67.1}
& \textbf{46.7}
& \textbf{65.3} \\

\midrule[1.0pt]

\textit{\textbf{Qwen3-32B}}
& 54.3
& 39.1
& 6.5
& \underline{55.3}
& 2.9
& 39.5
& 21.0
& 5.8
& 8.3
& 10.3
& 9.1
& \underline{84.1}
& \underline{64.3}
& \underline{39.3}
& \underline{40.9} \\

\hspace{6pt} $\bullet$ ReAct
& 58.9
& 33.8
& 34.0
& 40.5
& 10.4
& 18.4
& 26.2
& 9.0
& 12.5
& 8.8
& 7.7
& 72.5
& 55.9
& 34.8
& 27.0 \\

\hspace{6pt} $\bullet$ Plan-and-Solve
& 56.3
& \underline{39.7}
& 17.4
& 19.2
& 5.7
& 9.6
& 13.6
& 7.3
& 10.5
& 5.9
& 3.8
& 49.3
& 37.6
& 18.0
& 17.0 \\

\hspace{6pt} $\bullet$ Plan-and-Execute
& 58.9
& 31.9
& 13.1
& 12.6
& 9.7
& 7.5
& 2.2
& 1.8
& 2.8
& 2.0
& 3.0
& 47.8
& 33.4
& 21.9
& 14.4 \\

\hspace{6pt} $\bullet$ LawThinker
& 56.8
& \textbf{42.3}
& \underline{83.9}
& 53.5
& \textbf{86.7}
& \underline{42.0}
& 19.4
& 9.7
& \underline{19.6}
& \underline{17.5}
& \underline{79.8}
& 62.3
& 45.3
& 29.7
& 31.0 \\

\hspace{6pt} $\bullet$ Reflexion
& \underline{59.4}
& 31.4
& 43.7
& 38.2
& 25.4
& 29.0
& \underline{30.4}
& \underline{10.8}
& 16.6
& 12.3
& 6.6
& 58.0
& 43.9
& 30.1
& 19.9 \\

\rowcolor{gray!12}
\hspace{6pt} $\bullet$ \textbf{\method}
& \textbf{62.6}
& 39.4
& \textbf{100.0}
& \textbf{70.3}
& \underline{58.7}
& \textbf{53.5}
& \textbf{85.0}
& \textbf{23.3}
& \textbf{32.7}
& \textbf{31.4}
& \textbf{85.4}
& \textbf{98.6}
& \textbf{69.5}
& \textbf{45.7}
& \textbf{57.5} \\

\bottomrule
\end{tabular}
\end{threeparttable}
}

\end{table*}

\begin{table*}[!t]
\centering
\caption{Performance comparisons on J1-EVAL using InternLM3-8B and GLM-4-9B backbones. Best results are in \textbf{bold}, second-best results are \underline{underlined}, and ``-'' denotes failure to complete the task.}
\label{tab:main_result_other_J1Eval}

\resizebox{0.9\linewidth}{!}{
\begin{threeparttable}
\begin{tabular}{lccccccccccccccc}
\toprule

\multirow{3}{*}{\textbf{Method}}
& \multicolumn{2}{c}{\textbf{Level-I}}
& \multicolumn{4}{c}{\textbf{Level-II}}
& \multicolumn{9}{c}{\textbf{Level-III}} \\

\cmidrule(lr){2-3}
\cmidrule(lr){4-7}
\cmidrule(lr){8-16}

& \textbf{KQ}
& \textbf{LC}
& \multicolumn{2}{c}{\textbf{CD}}
& \multicolumn{2}{c}{\textbf{DD}}
& \multicolumn{4}{c}{\textbf{CI}}
& \multicolumn{5}{c}{\textbf{CR}} \\

\cmidrule(lr){2-2}
\cmidrule(lr){3-3}
\cmidrule(lr){4-5}
\cmidrule(lr){6-7}
\cmidrule(lr){8-11}
\cmidrule(lr){12-16}

& \textbf{Avg.}
& \textbf{Avg.}
& \textbf{FOR}
& \textbf{DOC}
& \textbf{FOR}
& \textbf{DOC}
& \textbf{PFS}
& \textbf{JUD}
& \textbf{REA}
& \textbf{LAW}
& \textbf{PFS}
& \textbf{CRI}
& \textbf{VER}
& \textbf{REA}
& \textbf{LAW} \\

\midrule[1.0pt]

\textit{\textbf{InternLM3-8B}}
& 39.8
& 33.0
& 49.0
& \underline{61.6}
& 1.4
& 27.3
& 5.7
& 1.5
& 2.3
& 1.4
& 8.6
& \underline{47.8}
& 36.3
& \underline{16.1}
& \underline{13.4} \\

\hspace{6pt} $\bullet$ ReAct
& 44.1
& 26.7
& 26.9
& 30.0
& -
& -
& -
& -
& -
& -
& -
& 2.9
& 1.0
& 0.7
& 0.4 \\

\hspace{6pt} $\bullet$ Plan-and-Solve
& 41.4
& 29.7
& 16.1
& 41.8
& -
& 5.8
& 3.8
& 3.1
& 8.5
& \underline{4.7}
& 2.0
& 37.7
& 25.2
& 12.3
& 9.9 \\

\hspace{6pt} $\bullet$ Plan-and-Execute
& \underline{47.6}
& 33.4
& 4.5
& 1.8
& -
& 0.4
& 4.5
& \underline{3.8}
& \underline{8.7}
& 3.9
& 0.0
& 5.8
& 5.2
& 1.3
& 1.5 \\

\hspace{6pt} $\bullet$ LawThinker
& 42.8
& \textbf{37.1}
& \underline{78.5}
& 60.1
& \underline{49.1}
& \underline{33.7}
& \underline{6.0}
& 2.7
& 3.9
& 2.1
& \underline{46.2}
& 26.1
& \underline{48.5}
& 6.2
& 8.1 \\

\hspace{6pt} $\bullet$ Reflexion
& 17.8
& 12.5
& 1.1
& 5.6
& -
& 0.3
& -
& -
& -
& -
& -
& -
& -
& -
& - \\

\rowcolor{gray!12}
\hspace{6pt} $\bullet$ \textbf{\method}
& \textbf{50.9}
& \underline{34.9}
& \textbf{100.0}
& \textbf{67.5}
& \textbf{56.3}
& \textbf{50.7}
& \textbf{88.9}
& \textbf{17.5}
& \textbf{21.9}
& \textbf{18.8}
& \textbf{78.2}
& \textbf{64.2}
& \textbf{61.7}
& \textbf{37.7}
& \textbf{20.4} \\

\midrule[1.0pt]

\textit{\textbf{GLM-4-9B}}
& 47.6
& 19.0
& 59.8
& \underline{31.1}
& 3.2
& \underline{13.5}
& 1.5
& -
& 0.2
& -
& \underline{15.0}
& \underline{40.6}
& \underline{26.2}
& \underline{12.6}
& \underline{8.4} \\

\hspace{6pt} $\bullet$ ReAct
& 48.3
& 16.8
& 5.4
& 19.3
& 1.8
& 6.0
& 0.1
& -
& -
& -
& 2.2
& 11.6
& 4.4
& 3.2
& 0.8 \\

\hspace{6pt} $\bullet$ Plan-and-Solve
& 19.1
& 6.7
& 0.6
& 0.5
& 0.7
& 3.3
& 0.6
& \underline{0.4}
& \underline{0.3}
& -
& 0.5
& 4.3
& 4.2
& 1.7
& 0.5 \\

\hspace{6pt} $\bullet$ Plan-and-Execute
& 35.6
& 12.1
& -
& -
& -
& -
& -
& -
& -
& -
& -
& -
& -
& -
& - \\

\hspace{6pt} $\bullet$ LawThinker
& \underline{53.9}
& \underline{24.6}
& \underline{63.2}
& 28.2
& \underline{6.8}
& 8.2
& \underline{72.0}
& -
& -
& -
& 3.6
& 15.9
& 9.1
& 4.6
& 3.8 \\

\hspace{6pt} $\bullet$ Reflexion
& 50.9
& 22.3
& 0.6
& 8.1
& 0.4
& 4.4
& -
& -
& -
& -
& 0.9
& 4.3
& 2.1
& 0.7
& 0.5 \\

\rowcolor{gray!12}
\hspace{6pt} $\bullet$ \textbf{\method}
& \textbf{55.6}
& \textbf{34.0}
& \textbf{95.5}
& \textbf{53.8}
& \textbf{57.8}
& \textbf{44.6}
& \textbf{80.0}
& \textbf{8.7}
& \textbf{12.0}
& \textbf{9.5}
& \textbf{72.5}
& \textbf{94.2}
& \textbf{51.6}
& \textbf{32.8}
& \textbf{22.0} \\

\bottomrule
\end{tabular}
\end{threeparttable}
}

\end{table*}


\textbf{Datasets}. To evaluate legal agents in long-horizon and multi-turn interaction settings, we consider two recent benchmarks, J1-EVAL~\citep{jia2026ready} and LegalWorld~\citep{zuo2026legalworld}, which cover diverse legal tasks ranging from interactive consultation to multi-stage litigation.

\textbf{J1-EVAL}~\citep{jia2026ready} contains six interactive legal scenarios covering different stages and roles in legal practice: Knowledge Questioning (KQ), Legal Consultation (LC), Complaint Drafting (CD), Defence Drafting (DD), Civil Court (CI), and Criminal Court (CR). Beyond static legal question answering, these tasks require agents to interact with simulated users or legal participants over multiple turns, progressively collect case-relevant information, generate legal documents, and follow scenario-specific procedures. The benchmark therefore evaluates both legal reasoning and the agent's ability to maintain coherent behavior throughout multi-turn interactions.

\begin{figure*}[t]
    \centering
    \includegraphics[width=\textwidth]{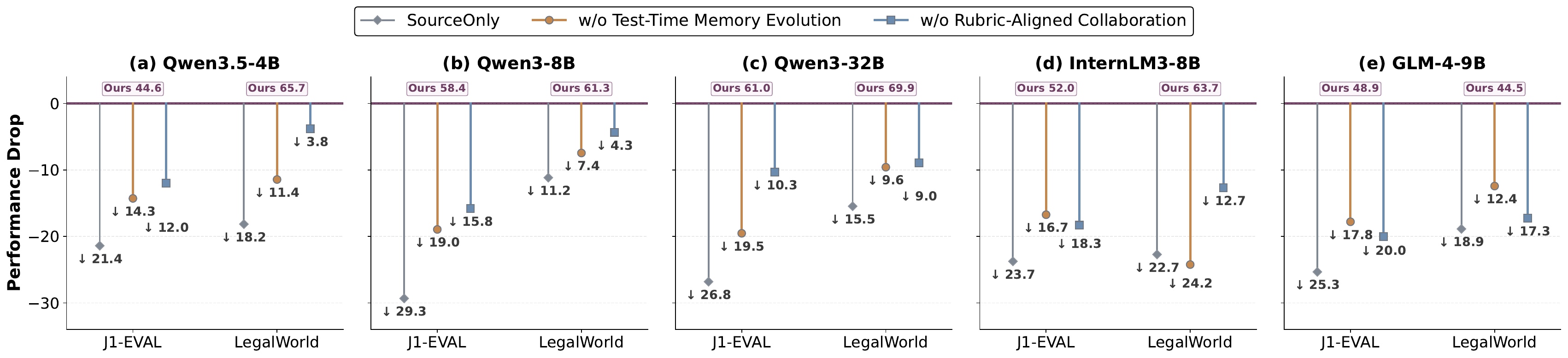}
\caption{
Ablation study of \method{} across five backbone models on J1-EVAL and LegalWorld.
Performance drops are reported after removing TME or RAC, showing that both components consistently contribute across models and benchmarks.
}
    \label{exp:ablation}
\end{figure*}

\begin{table}[!t]
\centering
\caption{Performance comparisons on LegalWorld using Qwen3.5-4B, Qwen3-8B, and Qwen3-32B backbones. Best results are in \textbf{bold} and second-best results are \underline{underlined}.}
\label{tab:legalworld_qwen}

\resizebox{\linewidth}{!}{
\begin{threeparttable}
\begin{tabular}{lcccccccc}
\toprule


\multirow{2}{*}{\textbf{Method}}
& \multicolumn{1}{c}{\textbf{LC}}
& \multicolumn{4}{c}{\textbf{DD}}
& \multicolumn{3}{c}{\textbf{CA}} \\

\cmidrule(lr){2-2}
\cmidrule(lr){3-6}
\cmidrule(lr){7-9}

& \textbf{IS}
& \textbf{PI}
& \textbf{CC}
& \textbf{FM}
& \textbf{EM}
& \textbf{PC}
& \textbf{EA}
& \textbf{LR} \\

\midrule[1.0pt]

\textit{\textbf{Qwen3.5-4B}}
& \underline{46.4}
& 56.8
& 53.5
& 49.7
& 33.4
& 51.8
& 42.6
& 45.9 \\

\hspace{6pt} $\bullet$ ReAct
& 43.7
& 57.4
& 55.4
& 50.5
& 40.2
& \underline{54.1}
& \underline{44.1}
& \underline{48.6} \\

\hspace{6pt} $\bullet$ Plan-and-Solve
& 41.8
& 55.7
& 46.4
& 41.8
& 27.1
& 45.2
& 36.4
& 41.3 \\

\hspace{6pt} $\bullet$ Plan-and-Execute
& 43.2
& 50.9
& 42.5
& 38.0
& 19.2
& 44.2
& 36.0
& 42.0 \\

\hspace{6pt} $\bullet$ LawThinker
& 43.4
& 51.9
& 54.8
& \underline{50.9}
& \underline{44.4}
& 40.1
& 31.7
& 38.3 \\

\hspace{6pt} $\bullet$ Reflexion
& 45.1
& \underline{58.8}
& \underline{55.8}
& 50.2
& 28.9
& 53.0
& 44.0
& 47.5 \\

\rowcolor{gray!12}
\hspace{6pt} $\bullet$ \textbf{\method}
& \textbf{53.5}
& \textbf{94.9}
& \textbf{68.4}
& \textbf{61.9}
& \textbf{65.8}
& \textbf{66.5}
& \textbf{55.8}
& \textbf{58.4} \\

\midrule[1.0pt]

\textit{\textbf{Qwen3-8B}}
& \underline{33.9}
& 80.1
& 62.9
& 49.6
& 43.1
& 52.0
& 39.4
& 39.9 \\

\hspace{6pt} $\bullet$ ReAct
& 28.9
& 82.5
& \underline{63.8}
& 52.4
& 45.5
& 52.1
& 40.5
& 42.0 \\

\hspace{6pt} $\bullet$ Plan-and-Solve
& 6.0
& 51.3
& 23.0
& 19.7
& 7.7
& 31.7
& 21.9
& 25.0 \\

\hspace{6pt} $\bullet$ Plan-and-Execute
& 4.6
& 49.9
& 24.8
& 19.8
& 8.2
& 29.6
& 21.7
& 25.4 \\

\hspace{6pt} $\bullet$ LawThinker
& 25.7
& \underline{88.5}
& \textbf{70.1}
& \underline{55.0}
& \underline{53.9}
& 37.8
& 25.6
& 30.5 \\

\hspace{6pt} $\bullet$ Reflexion
& 31.2
& 83.9
& 62.5
& 51.8
& 46.9
& \underline{53.8}
& \underline{40.9}
& \underline{42.2} \\

\rowcolor{gray!12}
\hspace{6pt} $\bullet$ \textbf{\method}
& \textbf{39.9}
& \textbf{94.1}
& \textbf{70.1}
& \textbf{58.0}
& \textbf{64.8}
& \textbf{63.4}
& \textbf{49.6}
& \textbf{50.2} \\

\midrule[1.0pt]

\textit{\textbf{Qwen3-32B}}
& 43.2
& 71.0
& 58.3
& 53.6
& 49.5
& 59.3
& 47.7
& 52.5 \\

\hspace{6pt} $\bullet$ ReAct
& 40.9
& 72.0
& \underline{66.1}
& \underline{59.2}
& \underline{58.4}
& \underline{60.3}
& \underline{49.2}
& \underline{52.8} \\

\hspace{6pt} $\bullet$ Plan-and-Solve
& 42.5
& 59.2
& 41.4
& 34.7
& 22.6
& 41.7
& 32.2
& 37.7 \\

\hspace{6pt} $\bullet$ Plan-and-Execute
& 40.1
& 60.8
& 36.2
& 30.7
& 17.4
& 36.1
& 27.5
& 32.5 \\

\hspace{6pt} $\bullet$ LawThinker
& \underline{47.5}
& \underline{76.1}
& 63.3
& 56.1
& 51.4
& 34.8
& 26.9
& 32.7 \\

\hspace{6pt} $\bullet$ Reflexion
& 43.3
& 75.0
& 64.9
& 56.6
& 49.9
& 58.4
& 47.3
& 51.5 \\

\rowcolor{gray!12}
\hspace{6pt} $\bullet$ \textbf{\method}
& \textbf{54.9}
& \textbf{95.5}
& \textbf{75.2}
& \textbf{69.5}
& \textbf{77.2}
& \textbf{69.4}
& \textbf{57.2}
& \textbf{60.0} \\

\bottomrule
\end{tabular}
\end{threeparttable}
}

\end{table}

\textbf{LegalWorld}~\citep{zuo2026legalworld} simulates a long-horizon civil litigation process through seven interconnected stages: Legal Consultation (LC), Complaint Drafting (CD), Defence Drafting (DD), First-Instance Trial (FIT), Appeal Drafting (AD), Appeal Response (AR), and Second-Instance Trial (SIT). In contrast to independently evaluated scenarios, the outputs and case states produced in earlier stages are carried forward to subsequent stages, forming a continuous litigation trajectory. This setting evaluates whether legal agents can preserve and utilize long-term case information, adapt to evolving procedural states, and maintain consistent reasoning across different stages of litigation.


\textbf{Baselines}. We compare our method with several representative reasoning and agent baselines. \textbf{SourceOnly} directly uses the original backbone model without additional design. We consider five representative open-source LLMs from three model families: GLM-4-9B~\citep{glm2024chatglm}, InternLM3-8B~\citep{cai2024internlm2}, Qwen3-8B, Qwen3-32B, and Qwen3.5-4B~\citep{qwen3technicalreport}. These models cover different model scales and serve as the base models for evaluating the effectiveness and generalizability of different agent frameworks.
\textbf{ReAct}~\citep{yao2022react} interleaves reasoning and action generation, allowing the agent to iteratively interact with external tools and update its reasoning based on environmental observations. \textbf{Plan-and-Solve}~\citep{wang2023plan} first decomposes a complex task into a sequence of manageable subtasks and then solves them sequentially according to the generated plan. \textbf{Plan-and-Execute}~\citep{topsakal2023creating} further separates high-level planning from step-wise execution, where a planner generates a global task plan and an executor carries out individual steps with available tools. \textbf{LawThinker}~\citep{yang2026lawthinker} is a legal-specific agent based on an Explore--Verify--Memorize workflow, which retrieves legal knowledge, explicitly verifies intermediate evidence and reasoning, and stores validated information for reuse in subsequent interactions. For a fair comparison, all methods are instantiated with the same backbone models and evaluation settings, and agent-based baselines are provided with the same set of external tools.


\textbf{Implementation Details}. Each backbone model is used to instantiate all agent roles. To avoid unbounded interaction loops, we follow the interaction settings of the original benchmarks~\citep{jia2026ready,zuo2026legalworld}. For LLM-based metric evaluation, we use DeepSeek-Flash~\citep{xu2026deepseek}. All experiments are conducted on NVIDIA L20 GPUs.

\begin{table}[!t]
\centering
\caption{Performance comparisons on LegalWorld using InternLM3-8B and GLM-4-9B backbones. Best results are in \textbf{bold} and second-best results are \underline{underlined}.}
\label{tab:legalworld_other}

\resizebox{\linewidth}{!}{
\begin{threeparttable}
\begin{tabular}{lcccccccc}
\toprule


\multirow{2}{*}{\textbf{Method}}
& \multicolumn{1}{c}{\textbf{LC}}
& \multicolumn{4}{c}{\textbf{DD}}
& \multicolumn{3}{c}{\textbf{CA}} \\

\cmidrule(lr){2-2}
\cmidrule(lr){3-6}
\cmidrule(lr){7-9}

& \textbf{IS}
& \textbf{PI}
& \textbf{CC}
& \textbf{FM}
& \textbf{EM}
& \textbf{PC}
& \textbf{EA}
& \textbf{LR} \\

\midrule[1.0pt]

\textit{\textbf{InternLM3-8B}}
& 41.1
& 67.5
& 62.4
& 50.8
& 41.0
& 23.6
& 18.3
& \underline{23.2} \\

\hspace{6pt} $\bullet$ ReAct
& 33.6
& \underline{78.0}
& \underline{69.9}
& 56.6
& 59.6
& 22.9
& 17.0
& 21.3 \\

\hspace{6pt} $\bullet$ Plan-and-Solve
& 34.9
& 58.9
& 43.2
& 35.9
& 38.5
& 22.2
& 16.2
& 20.2 \\

\hspace{6pt} $\bullet$ Plan-and-Execute
& 29.3
& 50.1
& 35.3
& 30.7
& 19.7
& \underline{26.7}
& \underline{19.7}
& \underline{23.7} \\

\hspace{6pt} $\bullet$ LawThinker
& \underline{42.3}
& 67.6
& 64.5
& 53.8
& \underline{60.0}
& 22.6
& 16.5
& 20.6 \\

\hspace{6pt} $\bullet$ Reflexion
& 34.7
& 72.7
& 68.4
& \underline{56.7}
& 56.0
& 23.4
& 18.3
& 21.9 \\

\rowcolor{gray!12}
\hspace{6pt} $\bullet$ \textbf{\method}
& \textbf{51.1}
& \textbf{92.3}
& \textbf{74.7}
& \textbf{61.2}
& \textbf{69.0}
& \textbf{63.7}
& \textbf{48.8}
& \textbf{48.9} \\

\midrule[1.0pt]

\textit{\textbf{GLM-4-9B}}
& \underline{29.4}
& 49.9
& 42.1
& 17.6
& 4.7
& \underline{24.5}
& \underline{17.7}
& \underline{18.6} \\

\hspace{6pt} $\bullet$ ReAct
& 25.7
& 54.3
& 44.2
& 15.4
& 7.6
& 18.0
& 13.1
& 13.7 \\

\hspace{6pt} $\bullet$ Plan-and-Solve
& 5.2
& 29.2
& 32.6
& 9.8
& 2.3
& 9.8
& 6.7
& 6.7 \\

\hspace{6pt} $\bullet$ Plan-and-Execute
& 1.0
& 24.2
& 22.6
& 9.8
& 5.0
& 9.4
& 5.7
& 5.5 \\

\hspace{6pt} $\bullet$ LawThinker
& 16.9
& \underline{57.4}
& \underline{45.8}
& \underline{20.8}
& \underline{15.2}
& 20.3
& 13.4
& 16.7 \\

\hspace{6pt} $\bullet$ Reflexion
& 25.5
& 56.3
& 44.2
& 14.1
& 6.4
& 20.9
& 14.9
& 16.2 \\

\rowcolor{gray!12}
\hspace{6pt} $\bullet$ \textbf{\method}
& \textbf{34.0}
& \textbf{84.6}
& \textbf{51.7}
& \textbf{28.4}
& \textbf{36.0}
& \textbf{49.9}
& \textbf{36.1}
& \textbf{35.1} \\

\bottomrule
\end{tabular}
\end{threeparttable}
}

\end{table}

\section{Results and Analysis}

\begin{figure*}[t]
    \centering
    \includegraphics[width=\textwidth]{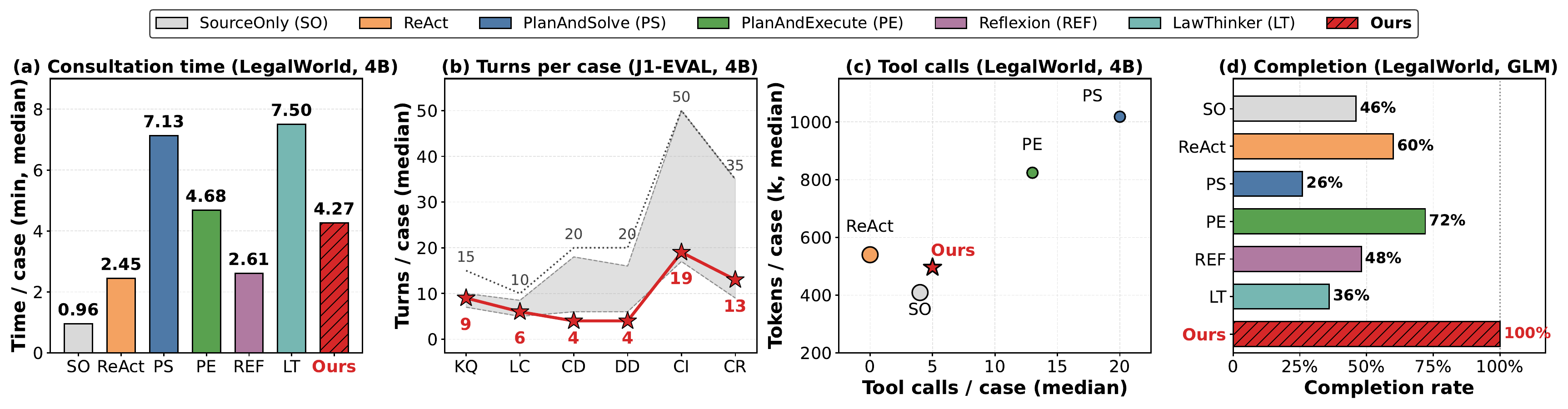}
    \caption{
    Efficiency analysis of \method.
    (a) Median execution time per case.
    (b) Median interaction turns across different task scenarios.
    (c) Tool-call and token usage per case.
    (d) End-to-end task completion rate.
    }
    \label{exp:effiency}
\end{figure*}

\subsection{Main Results}
As shown in Tables~\ref{tab:main_result_J1Eval} and~\ref{tab:main_result_other_J1Eval}, \method{} consistently improves performance on J1-EVAL across different backbone models and task levels. The gains are particularly pronounced on document drafting and courtroom-oriented metrics. Improvements on \textsc{FOR} and \textsc{DOC} indicate stronger preservation of task-specific information and more reliable generation of structurally complete legal documents. Meanwhile, the gains on \textsc{PFS}, \textsc{REA}, and \textsc{LAW} suggest better procedural compliance, legal reasoning, and use of appropriate legal provisions. 

Tables~\ref{tab:legalworld_qwen} and~\ref{tab:legalworld_other} further show that \method{} achieves consistent improvements on LegalWorld across legal consultation, document drafting, and courtroom advocacy. The improvements are especially evident in drafting and advocacy tasks, where successful completion requires agents to preserve case facts across stages, adapt to changing role objectives, and reuse previously established evidence. 


Across Tables~\ref{tab:main_result_J1Eval}, \ref{tab:main_result_other_J1Eval}, \ref{tab:legalworld_qwen}, and~\ref{tab:legalworld_other}, the improvements are consistently observed on Qwen3.5-4B, Qwen3-8B, Qwen3-32B, InternLM3-8B, and GLM-4-9B. This indicates that \method{} is not tied to a particular model family or parameter scale, but provides a generally effective test-time adaptation mechanism for improving long-horizon legal agents without updating the underlying model parameters.

\subsection{Ablation Study}
As shown in Fig.~\ref{exp:ablation}, removing either component consistently degrades the performance of \method{} across all five backbone models and both benchmarks. On J1-EVAL, removing Test-Time Memory Evolution leads to an average performance drop of 17.5 points, while removing Rubric-Aligned Collaboration results in a 15.3-point drop. Similar trends are observed on LegalWorld, with average decreases of 13.0 and 9.4 points after removing TME and RAC, respectively. Averaged over all model--benchmark combinations, the performance drops are 15.2 points for w/o TME and 12.4 points for w/o RAC. In comparison, SourceOnly exhibits an average degradation of 21.3 points relative to the full model. These results demonstrate that both components make substantial and complementary contributions: TME provides persistent task- and case-specific knowledge adaptation, whereas RAC regulates role-specific behavior and interaction progression; combining them yields the strongest overall performance.

\begin{table}[t]
\centering
\caption{Audit results for the self-judgment mechanisms in TME and RAC. 
All values are reported on a 0--100 scale.}
\label{tab:jev_audit}
\resizebox{0.8\linewidth}{!}{
\begin{tabular}{lcc}
\toprule
\textbf{Audit Metric} & \textbf{TME} & \textbf{RAC} \\
\midrule
Judgment Agreement              & 66.8 & 71.7 \\
Judge-Verified Local Error      & 32.6 & 39.6 \\
Jev Coverage at Matched Error   & 84.3 & 91.6 \\
Potential Local LLM Calls Saved & --   & 89.5 \\
\bottomrule
\end{tabular}
}
\end{table}

\begin{figure*}[!ht]
    \centering
    \includegraphics[width=\textwidth]{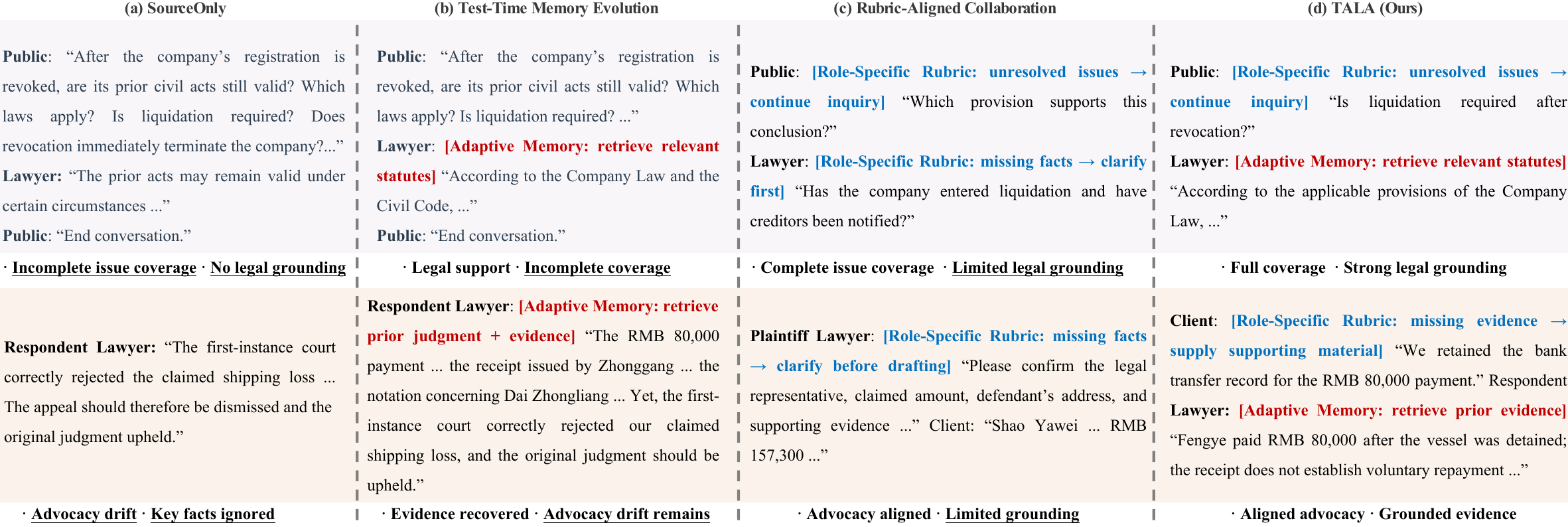}
    \caption{
    Case studies on J1-EVAL and LegalWorld.
    The upper case presents a legal consultation scenario in J1-EVAL, while the lower case illustrates a multi-stage legal workflow in LegalWorld.
    We compare SourceOnly with Test-Time Memory Evolution, Rubric-Aligned Collaboration, and the full TALA framework to show their complementary effects on legal grounding, interaction coverage, role alignment, and cross-stage evidence use.
    The takeaway beneath each case summarizes the observed behavior, where \underline{underlined} terms indicate remaining limitations.
    }
    \label{fig:case_study}
\end{figure*}

\subsection{Efficiency of \method}
We further evaluate the interaction and computational efficiency of \method{}. As shown in Fig.~\ref{exp:effiency}, \method{} maintains reasonable test-time overhead while supporting reliable long-horizon execution.

\textbf{1) Execution Time.}
On LegalWorld, \method{} requires a median of 4.27 minutes per consultation case, remaining below several planning-intensive baselines. This suggests that the additional adaptation steps introduce manageable computational overhead. \textbf{2) Interaction Turns.}
Across J1-EVAL scenarios, \method{} remains within the interaction range of existing methods. Its performance gains therefore do not simply result from extending the conversation with additional turns.
\textbf{3) Tool and Token Usage.}
\method{} maintains moderate tool-call and token usage by selectively invoking retrieval when additional legal knowledge or prior experience is needed. This avoids repeatedly expanding intermediate reasoning or invoking tools throughout the trajectory.
\textbf{4) Completion Rate.}
On LegalWorld with GLM-4-9B, \method{} achieves a 100\% completion rate, compared with 26\%--72\% for the evaluated baselines. This indicates improved robustness in sustaining long-horizon execution across multiple stages.

\subsection{Case Study Analysis}
Fig.~\ref{fig:case_study} further illustrates how the two components of \method{} affect agent behavior through representative cases. 
In the J1-EVAL consultation case, the user raises multiple legal issues concerning the validity of prior civil acts, applicable provisions, liquidation, and company termination. 
SourceOnly provides only a partial response and terminates the conversation before the remaining issues are addressed. 
\textit{TME} retrieves relevant legal provisions and improves the grounding of the lawyer's response, but does not prevent premature termination. 
In contrast, \textit{RAC} explicitly tracks unresolved issues and encourages the involved roles to continue clarification and inquiry, resulting in more complete issue coverage, although the available legal grounding remains limited. 
Combining both mechanisms, \method{} maintains the interaction until the required issues are covered while grounding the lawyer's responses with relevant legal knowledge.

The LegalWorld case demonstrates a different failure mode in a long-horizon, multi-stage legal workflow. 
The same dispute evolves from consultation and complaint drafting to first-instance and appellate proceedings, requiring the agent to preserve case evidence while adapting to changing role objectives. 
SourceOnly exhibits advocacy drift: although acting as the respondent's lawyer, it focuses on defending the rejection of its own claimed shipping loss rather than protecting the favorable RMB~80,000 judgment. 
Test-Time Memory Evolution successfully recovers prior-stage evidence, including the payment and receipt information, but the advocacy objective can still remain misaligned. 
Rubric-Aligned Collaboration instead regulates stage- and role-specific behavior, such as requiring missing facts to be clarified before drafting, but cannot recover evidence that is unavailable in the current interaction. 
With both components, \method{} preserves relevant evidence across stages while maintaining role-aligned advocacy.

Together, the two cases highlight the complementary roles of the proposed components. 
Test-Time Memory Evolution primarily determines \emph{what information should be recovered and reused}, whereas Rubric-Aligned Collaboration regulates \emph{how each role should act and how the interaction should progress}. 
Their combination therefore addresses both knowledge continuity and behavioral alignment, which are jointly required for reliable long-horizon agent execution.

\subsection{Auditing Self-Judgment with Jev}
We further audit the self-judgment mechanisms in \method{} using Jev\footnote{\url{https://typesafe.ai/}}, an independent lightweight structured judge. As shown in Table~\ref{tab:jev_audit}, we conduct the analysis on the J1-EVAL KQ scenario across all five backbone models, covering the internal judgments used by both Test-Time Memory Evolution (TME) and Rubric-Aligned Collaboration (RAC). We report four metrics: \emph{Judgment Agreement} measures the consistency between Jev and the original local self-judgment; \emph{Judge-Verified Local Error} estimates the error rate of the original judgment through independent verification; \emph{Jev Coverage at Matched Error} measures the proportion of decisions that Jev can handle without exceeding the original judgment error; and \emph{Potential Local LLM Calls Saved} estimates the corresponding reduction in local LLM-based judgment calls. Overall, the results support the effectiveness of our internal judgment design, while lightweight judges such as Jev provide a promising direction for further improving verification efficiency and reducing local LLM calls.

\section{Conclusion}
In this paper, we investigate test-time evolution for long-horizon legal agents and identify two key challenges in realistic deployment: \emph{case heterogeneity}, which complicates experience transfer across continuously arriving cases, and \emph{cross-role coupling}, where local decisions can influence subsequent roles and propagate across procedural stages. To address these challenges, we propose \method, which enables agents to continually adapt throughout deployment by evolving cross-case experience and coordinating decisions across coupled legal roles and stages. Specifically, \emph{Test-Time Memory Evolution} retrieves, adapts, and consolidates prior interaction experience to support heterogeneous cases, while \emph{Rubric-Aligned Collaboration} aligns role decisions with behavioral and procedural requirements to improve global consistency across long-horizon interactions. Extensive experiments on J1-EVAL and LegalWorld across five backbone models demonstrate consistent improvements across model families and scales, together with stronger robustness in multi-stage execution. Further ablation, efficiency, and case studies show that the two components are complementary and mutually reinforcing, jointly improving cross-case adaptation, cross-role coordination, and the overall reliability of long-horizon legal reasoning.

\textbf{Limitations.}
Our current study focuses on autonomous test-time evolution, while real-world legal decision-making is inherently interactive and often requires sustained collaboration between agents and human professionals. Future work may extend \method{} toward human-in-the-loop and recursively interactive legal agents that continuously incorporate expert feedback, user corrections, and procedural changes during deployment. Evaluating such systems in more specialized legal settings, with stronger requirements on accountability, professional expertise, and human-agent coordination, remains an important direction.




\bibliographystyle{ACM-Reference-Format}
\bibliography{references}

@inproceedings{jia2026ready,
  title={Ready jurist one: Benchmarking language agents for legal intelligence in dynamic environments},
  author={Jia, Zheng and Yue, Shengbin and Chen, Wei and Wang, Siyuan and Liu, Yidong and Li, Zejun and Song, Yun and Wei, Zhongyu},
  booktitle={Proceedings of the 64th Annual Meeting of the Association for Computational Linguistics (Volume 1: Long Papers)},
  pages={10351--10376},
  year={2026}
}

@article{zuo2026legalworld,
  title={LegalWorld: A Life-Cycle Interactive Environment for Legal Agents},
  author={Zuo, Songhan and Yue, Shengbin and Chiang, Tao and Li, Guanying and Song, Yun and Huang, Xuanjing and Wei, Zhongyu},
  journal={arXiv preprint arXiv:2606.18728},
  year={2026}
}

@article{yao2022react,
  title={React: Synergizing reasoning and acting in language models},
  author={Yao, Shunyu and Zhao, Jeffrey and Yu, Dian and Du, Nan and Shafran, Izhak and Narasimhan, Karthik and Cao, Yuan},
  journal={arXiv preprint arXiv:2210.03629},
  year={2022}
}

@inproceedings{wang2023plan,
  title={Plan-and-solve prompting: Improving zero-shot chain-of-thought reasoning by large language models},
  author={Wang, Lei and Xu, Wanyu and Lan, Yihuai and Hu, Zhiqiang and Lan, Yunshi and Lee, Roy Ka-Wei and Lim, Ee-Peng},
  booktitle={Proceedings of the 61st annual meeting of the association for computational linguistics (volume 1: Long papers)},
  pages={2609--2634},
  year={2023}
}

@inproceedings{topsakal2023creating,
  title={Creating large language model applications utilizing langchain: A primer on developing llm apps fast},
  author={Topsakal, Oguzhan and Akinci, Tahir Cetin},
  booktitle={International conference on applied engineering and natural sciences},
  volume={1},
  number={1},
  pages={1050--1056},
  year={2023}
}

@article{yang2026lawthinker,
  title={Lawthinker: A deep research legal agent in dynamic environments},
  author={Yang, Xinyu and Deng, Chenlong and Wen, Tongyu and Xie, Binyu and Dou, Zhicheng},
  journal={arXiv preprint arXiv:2602.12056},
  year={2026}
}

@misc{qwen3technicalreport,
      title={Qwen3 Technical Report}, 
      author={Qwen Team},
      year={2025},
      eprint={2505.09388},
      archivePrefix={arXiv},
      primaryClass={cs.CL},
      url={https://arxiv.org/abs/2505.09388}, 
}

@article{glm2024chatglm,
  title={Chatglm: A family of large language models from glm-130b to glm-4 all tools},
  author={Glm, Team and Zeng, Aohan and Xu, Bin and Wang, Bowen and Zhang, Chenhui and Yin, Da and Zhang, Dan and Rojas, Diego and Feng, Guanyu and Zhao, Hanlin and others},
  journal={arXiv preprint arXiv:2406.12793},
  year={2024}
}

@article{cai2024internlm2,
  title={Internlm2 technical report},
  author={Cai, Zheng and Cao, Maosong and Chen, Haojiong and Chen, Kai and Chen, Keyu and Chen, Xin and Chen, Xun and Chen, Zehui and Chen, Zhi and Chu, Pei and others},
  journal={arXiv preprint arXiv:2403.17297},
  year={2024}
}

@inproceedings{liu2026llm,
  title={LLM agents in law: Taxonomy, applications, and challenges},
  author={Liu, Shuang and Zhang, Ruijia and Ma, Ruoyun and Deng, Yujia and Zhu, Lanyi and Li, Jiayu and Li, Zelong and Shen, Zhibin and Du, Mengnan},
  booktitle={Proceedings of the 64th Annual Meeting of the Association for Computational Linguistics (Volume 1: Long Papers)},
  pages={15768--15792},
  year={2026}
}

@inproceedings{patil2025berkeley,
  title={The berkeley function calling leaderboard (bfcl): From tool use to agentic evaluation of large language models},
  author={Patil, Shishir G and Mao, Huanzhi and Yan, Fanjia and Ji, Charlie Cheng-Jie and Suresh, Vishnu and Stoica, Ion and Gonzalez, Joseph E},
  booktitle={Forty-second International Conference on Machine Learning},
  year={2025}
}

@misc{bge_embedding,
      title={C-Pack: Packaged Resources To Advance General Chinese Embedding}, 
      author={Shitao Xiao and Zheng Liu and Peitian Zhang and Niklas Muennighoff},
      year={2023},
      eprint={2309.07597},
      archivePrefix={arXiv},
      primaryClass={cs.CL}
}

@inproceedings{cao2026higmem,
  title={HiGMem: A Hierarchical and LLM-Guided Memory System for Long-Term Conversational Agents},
  author={Cao, Shuqi and He, Jingyi and Tan, Fei},
  booktitle={Findings of the Association for Computational Linguistics: ACL 2026},
  pages={33853--33862},
  year={2026}
}

@inproceedings{yue2025multi,
  title={Multi-agent simulator drives language models for legal intensive interaction},
  author={Yue, Shengbin and Huang, Ting and Jia, Zheng and Wang, Siyuan and Liu, Shujun and Song, Yun and Huang, Xuan-Jing and Wei, Zhongyu},
  booktitle={Findings of the Association for Computational Linguistics: NAACL 2025},
  pages={6552--6585},
  year={2025}
}

@inproceedings{wu2024knowledge,
  title={Knowledge-infused legal wisdom: Navigating llm consultation through the lens of diagnostics and positive-unlabeled reinforcement learning},
  author={Wu, Yang and Wang, Chenghao and Gumusel, Ece and Liu, Xiaozhong},
  booktitle={Findings of the Association for Computational Linguistics: ACL 2024},
  pages={15542--15555},
  year={2024}
}

@inproceedings{fei2024lawbench,
  title={Lawbench: Benchmarking legal knowledge of large language models},
  author={Fei, Zhiwei and Shen, Xiaoyu and Zhu, Dawei and Zhou, Fengzhe and Han, Zhuo and Huang, Alan and Zhang, Songyang and Chen, Kai and Yin, Zhixin and Shen, Zongwen and others},
  booktitle={Proceedings of the 2024 conference on empirical methods in natural language processing},
  pages={7933--7962},
  year={2024}
}

@article{yue2023disc,
  title={Disc-lawllm: Fine-tuning large language models for intelligent legal services},
  author={Yue, Shengbin and Chen, Wei and Wang, Siyuan and Li, Bingxuan and Shen, Chenchen and Liu, Shujun and Zhou, Yuxuan and Xiao, Yao and Yun, Song and Huang, Xuanjing and others},
  journal={arXiv preprint arXiv:2309.11325},
  year={2023}
}

@inproceedings{fei2025internlm,
  title={Internlm-law: An open-sourced chinese legal large language model},
  author={Fei, Zhiwei and Zhang, Songyang and Shen, Xiaoyu and Zhu, Dawei and Wang, Xiao and Ge, Jidong and Ng, Vincent},
  booktitle={Proceedings of the 31st International Conference on Computational Linguistics},
  pages={9376--9392},
  year={2025}
}

@article{colombo2024saullm,
  title={Saullm-7b: A pioneering large language model for law},
  author={Colombo, Pierre and Pires, Telmo Pessoa and Boudiaf, Malik and Culver, Dominic and Melo, Rui and Corro, Caio and Martins, Andre FT and Esposito, Fabrizio and Raposo, Vera L{\'u}cia and Morgado, Sofia and others},
  journal={arXiv preprint arXiv:2403.03883},
  year={2024}
}

@article{chen2025lexpro,
  title={LexPro-1.0 technical report},
  author={Chen, Haotian and Xu, Yanyu and Wang, Boyan and Zhao, Chaoyue and Han, Xiaoyu and Wang, Fang and Cui, Lizhen and Xu, Yonghui},
  journal={arXiv preprint arXiv:2503.06949},
  year={2025}
}

@article{cui2026chatlaw,
  title={Chatlaw: A multi-agent legal assistant based on a role-aligned mixture-of-experts architecture},
  author={Cui, Jiaxi and Ning, Munan and Li, Zongjian and Li, Hao and Ya, Yang and Chen, Bohua and Ling, Bin and Tian, Yonghong and Yuan, Li},
  journal={Fundamental Research},
  year={2026},
  publisher={Elsevier}
}

@inproceedings{chen2025agentcourt,
  title={Agentcourt: Simulating court with adversarial evolvable lawyer agents},
  author={Chen, Guhong and Fan, Liyang and Gong, Zihan and Xie, Nan and Li, Zixuan and Liu, Ziqiang and Li, Chengming and Qu, Qiang and Alinejad-Rokny, Hamid and Ni, Shiwen and others},
  booktitle={Findings of the Association for Computational Linguistics: ACL 2025},
  pages={5850--5865},
  year={2025}
}

@inproceedings{cai2025unilaw,
  title={Unilaw-r1: A large language model for legal reasoning with reinforcement learning and iterative inference},
  author={Cai, Hua and Zhao, Shuang and Zhang, Liang and Shen, Xuli and Xu, Qing and Shen, Weilin and Wen, Zihao and Ban, Tianke},
  booktitle={Proceedings of the 2025 Conference on Empirical Methods in Natural Language Processing},
  pages={18128--18142},
  year={2025}
}

@article{chen2026test,
  title={Test-Time Thinking: Reinforcement Adaptation for Multimodal LLM Reasoning},
  author={Chen, Haotian and Xu, Yanyu and Xu, Yonghui and Wang, Boyan and Wang, Fang and Cui, Lizhen},
  journal={Expert Systems with Applications},
  pages={132524},
  year={2026},
  publisher={Elsevier}
}

@inproceedings{huang2026sample,
  title={Sample complexity and representation ability of test-time scaling paradigms},
  author={Huang, Baihe and Li, Shanda and Wu, Tianhao and Yang, Yiming and Talwalkar, Ameet and Ramchandran, Kannan and Jordan, Michael and Jiao, Jiantao},
  booktitle={International Conference on Learning Representations},
  volume={2026},
  pages={54262--54298},
  year={2026}
}

@article{wang2022self,
  title={Self-consistency improves chain of thought reasoning in language models},
  author={Wang, Xuezhi and Wei, Jason and Schuurmans, Dale and Le, Quoc and Chi, Ed and Narang, Sharan and Chowdhery, Aakanksha and Zhou, Denny},
  journal={arXiv preprint arXiv:2203.11171},
  year={2022}
}

@article{madaan2023self,
  title={Self-refine: Iterative refinement with self-feedback},
  author={Madaan, Aman and Tandon, Niket and Gupta, Prakhar and Hallinan, Skyler and Gao, Luyu and Wiegreffe, Sarah and Alon, Uri and Dziri, Nouha and Prabhumoye, Shrimai and Yang, Yiming and others},
  journal={Advances in neural information processing systems},
  volume={36},
  pages={46534--46594},
  year={2023}
}

@inproceedings{snell2025scaling,
  title={Scaling LLM test-time compute optimally can be more effective than scaling parameters for reasoning},
  author={Snell, Charlie and Lee, Jaehoon and Xu, Kelvin and Kumar, Aviral},
  booktitle={International Conference on Learning Representations},
  volume={2025},
  pages={10131--10165},
  year={2025}
}

@article{huang2025best,
  title={Is best-of-n the best of them? coverage, scaling, and optimality in inference-time alignment},
  author={Huang, Audrey and Block, Adam and Liu, Qinghua and Jiang, Nan and Krishnamurthy, Akshay and Foster, Dylan J},
  journal={arXiv preprint arXiv:2503.21878},
  year={2025}
}

@inproceedings{enguehard2025lemaj,
  title={Lemaj (legal llm-as-a-judge): Bridging legal reasoning and llm evaluation},
  author={Enguehard, Joseph and Van Ermengem, Morgane and Atkinson, Kate and Cha, Sujeong and Chowdhury, Arijit Ghosh and Ramaswamy, Prashanth Kallur and Roghair, Jeremy and Marlowe, Hannah R and Negreanu, Carina Suzana and Boxall, Kitty and others},
  booktitle={Proceedings of the natural legal language processing workshop 2025},
  pages={318--337},
  year={2025}
}

@inproceedings{li2025legalagentbench,
  title={Legalagentbench: Evaluating llm agents in legal domain},
  author={Li, Haitao and Chen, Junjie and Yang, Jingli and Ai, Qingyao and Jia, Wei and Liu, Youfeng and Lin, Kai and Wu, Yueyue and Yuan, Guozhi and Hu, Yiran and others},
  booktitle={Proceedings of the 63rd annual meeting of the association for computational linguistics (volume 1: long papers)},
  pages={2322--2344},
  year={2025}
}

@inproceedings{zhang2026defgen,
  title={DefGen-Bench: A Benchmark for Chinese Criminal Defence Opinion Generation in LegalAI},
  author={Zhang, Senbo and Wang, Qiqi and Lou, Fanghao and Chen, Guanyu and Pan, Yihong and Li, Huijia and Liu, Qian},
  booktitle={Proceedings of the 64th Annual Meeting of the Association for Computational Linguistics (Volume 1: Long Papers)},
  pages={35378--35392},
  year={2026}
}

@inproceedings{suzgun2026dynamic,
  title={Dynamic cheatsheet: Test-time learning with adaptive memory},
  author={Suzgun, Mirac and Yuksekgonul, Mert and Bianchi, Federico and Jurafsky, Dan and Zou, James},
  booktitle={Proceedings of the 19th Conference of the European Chapter of the Association for Computational Linguistics (Volume 1: Long Papers)},
  pages={7080--7106},
  year={2026}
}

@article{xu2026deepseek,
  title={Deepseek-v4: Towards highly efficient million-token context intelligence},
  author={Xu, Anyi and Lin, Bangcai and Xue, Bing and Wang, Bingxuan and Xu, Bingzheng and Wu, Bochao and Zhang, Bowei and Lin, Chaofan and Dong, Chen and Ling, Chenchen and others},
  journal={arXiv preprint arXiv:2606.19348},
  year={2026}
}

\appendix

\end{document}